\documentclass{article}

\PassOptionsToPackage{numbers, compress}{natbib}

\usepackage{amsmath,amsfonts,bm}

\def\eqref#1{equation~\ref{#1}}

\def\1{\bm{1}}

\DeclareMathAlphabet{\mathsfit}{\encodingdefault}{\sfdefault}{m}{sl}
\SetMathAlphabet{\mathsfit}{bold}{\encodingdefault}{\sfdefault}{bx}{n}

\usepackage{tikz}
\usetikzlibrary{matrix}

\usepackage{hyperref}
\usepackage{url}
\usepackage{graphicx}
\usepackage{subcaption}
\usepackage{multirow}
\usepackage{booktabs}       
\usepackage{wrapfig}
\usepackage{caption}
\usepackage{amsmath}
\usepackage{amssymb}
\usepackage{mathtools}
\usepackage{amsthm}
\usepackage{ddphonism}
\usepackage{hyperref}

\usepackage{amsmath}
\usepackage{amssymb}
\usepackage{mathtools}
\usepackage{amsthm}

\usepackage[capitalize,noabbrev]{cleveref}

\theoremstyle{plain}

\theoremstyle{definition}

\theoremstyle{remark}

\usepackage[textsize=tiny]{todonotes}

    \usepackage[preprint]{neurips_2024}

\usepackage[utf8]{inputenc} 
\usepackage[T1]{fontenc}    
\usepackage{hyperref}       
\usepackage{url}            
\usepackage{booktabs}       
\usepackage{amsfonts}       
\usepackage{nicefrac}       
\usepackage{microtype}      
\usepackage{xcolor}         

\title{Cyclic Sparse Training: Is it Enough?}

\author{%
  Advait Gadhikar, Sree Harsha Nelaturu \& Rebekka Burkholz\\
  CISPA Helmholtz Center for Information Security\\
  Saarbrücken, Germany, 66123\\
  \texttt{\{advait.gadhikar, harsha.nelaturu, burkholz\}@cispa.de}\\
}

\begin{document}

\maketitle

\begin{abstract}

The success of iterative pruning methods in achieving state-of-the-art sparse networks has largely been attributed to improved mask identification and an implicit regularization induced by pruning. We challenge this hypothesis and instead posit that their repeated cyclic training schedules enable improved optimization. To verify this, we show that pruning at initialization is significantly boosted by repeated cyclic training, even outperforming standard iterative pruning methods. The dominant mechanism how this is achieved, as we conjecture, can be attributed to a better exploration of the loss landscape leading to a lower training loss. However, at high sparsity, repeated cyclic training alone is not enough for competitive performance. A strong coupling between learnt parameter initialization and mask seems to be required. Standard methods obtain this coupling via expensive pruning-training iterations, starting from a dense network. To achieve this with sparse training instead, we propose SCULPT-ing, i.e., repeated cyclic training of any sparse mask followed by a single pruning step to couple the parameters and the mask, which is able to match the performance of state-of-the-art iterative pruning methods in the high sparsity regime at reduced computational cost. 

\end{abstract}


\section{Introduction}

Overparameterization has been a key factor in the tremendous success of deep neural networks across a variety of tasks on vision and language \citep{bubeck2023sparks}, among others.
However, the massive model sizes come with the burden of high computational and memory costs \cite{wu2022sustainable,luccioni2023power}.
Hence, to ensure long-term benefits of deep learning for society and climate, it is imperative to improve model efficiency not only at inference time but also during training \cite{kaack2022aligning}.

Neural network sparsification offers a means to reduce the number of parameters of a model while minimally affecting its performance. 
In addition to computational and memory savings, it can also improve generalization \cite{frankle2019lottery,paul2023unmasking} and interpretability \cite{chen2022can,hossain2024tickets}, perform denoising \cite{prune-regularize,wang2023searching}, and introduce verifiability \cite{Narodytska2020In,albarghouthi2021introduction}. 
While state-of-the-art iterative pruning methods like Learning Rate Rewinding (LRR) \citep{rewindVsFinetune} or Iterative Magnitude Pruning (IMP) \citep{frankle2019lottery} are able to obtain highly performant sparse networks, they require training a dense network over multiple pruning and training iterations, which are computationally demanding. 

Instead, pruning at initialization (PaI) methods find a sparse mask at initialization that defines which parameters are pruned i.e. frozen to zero. 
It thus realizes computational and memory savings from the beginning of model development.  
While they aim to solve one of our most pressing problems by enabling sparse training from scratch,
these methods struggle to keep up with the performance of iterative pruning and often fall short at high sparsities, especially on more complex tasks \citep{frankle2021review}.

Why is this the case?
Recent work has attributed the success of iterative pruning methods to their ability to find better sparse masks \citep{paul2023unmasking}, to train flexibly by enabling more parameter sign flips \citep{lrr-signs,zhou2019deconstructing}, and to identify better trainable parameter initializations of the mask \citep{frankle2021review,kuznedelev2023accurate}. 
With the goal to fill the gap between PaI and iterative methods, we investigate to which degree we can transfer successful mechanisms from LRR, regarding its training procedure and mask learning ability, to achieve state-of-the-art performance with PaI methods.

\begin{figure}
\centering
\begin{subfigure}{.45\textwidth}
  \centering
  \includegraphics[width=\textwidth]{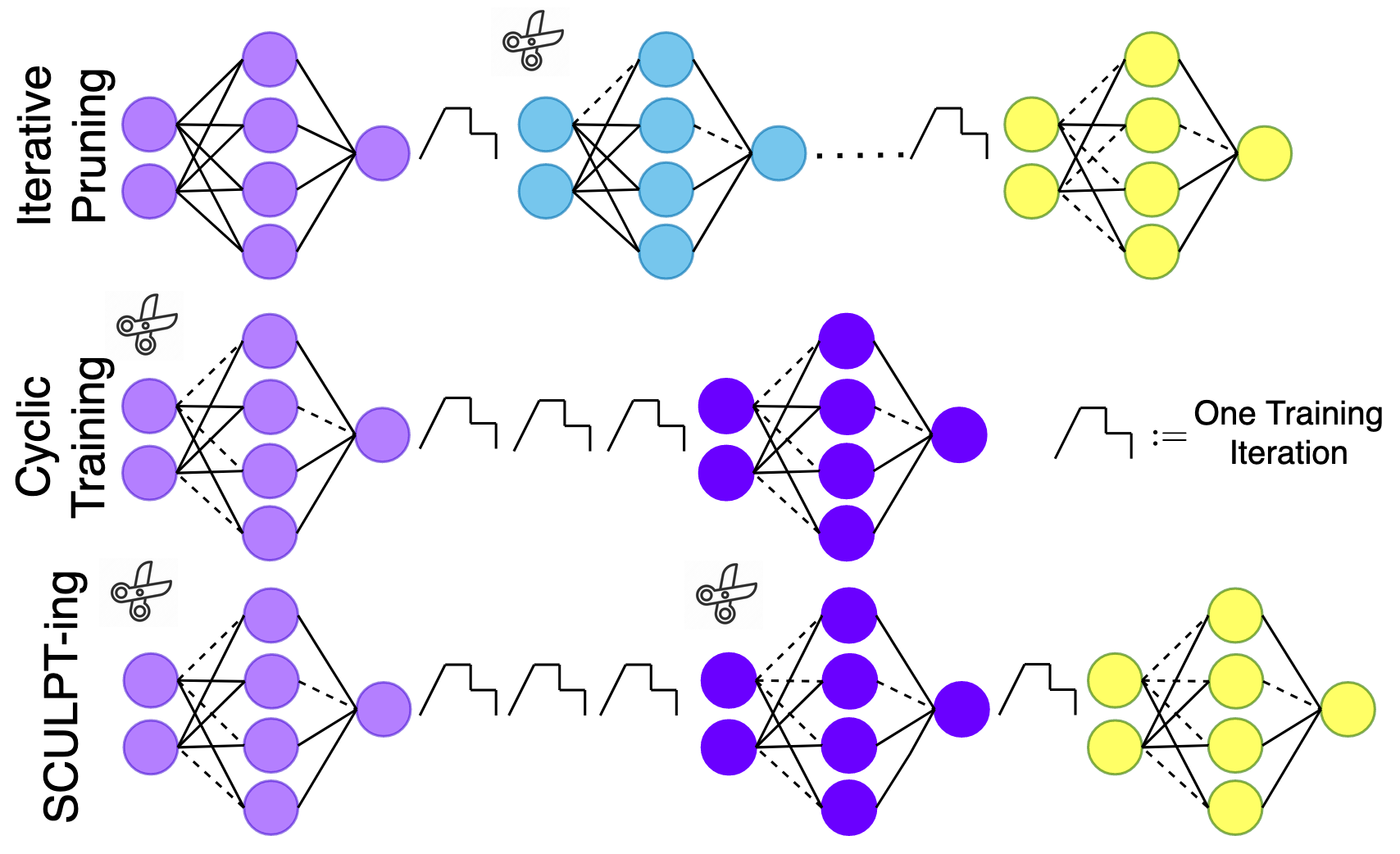}
  \caption{}
  \label{fig:cyclic-train}
\end{subfigure}%
\hfill
\begin{subfigure}{.48\textwidth}
  \centering
  \includegraphics[width=
\textwidth]{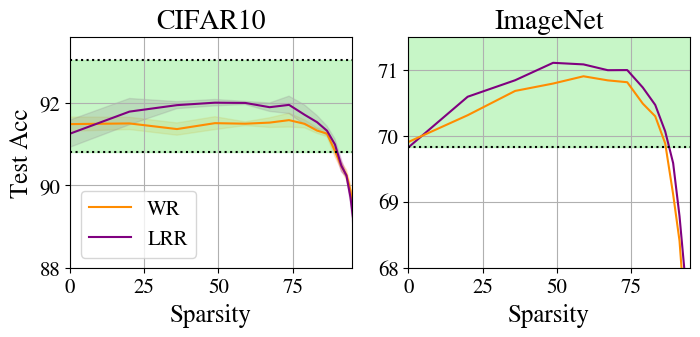}
\vspace{0.1pt}
  \caption{}
  \label{fig:lrr-peak}
\end{subfigure}
\caption{(a) Illustration of iterative pruning (top), cyclic training (middle) and SCULPT-ing (bottom). (b) Iterative pruning improves generalization on  CIFAR10 (left) and ImageNet (right). Shaded area denotes gain in performance for dense networks with cyclic training.}
\end{figure}

First, we study how LRR achieves peak performance in the sparse regime surpassing the performance of its dense counterpart (see Figure ~\ref{fig:lrr-peak}). 
This typical observation has been attributed to a sparsity induced regularization effect \citep{frankle2019lottery,han2015learning,prune-regularize}.
We offer an alternative explanation.
Instead of reaching an optimal sparsity level, we posit that the peak corresponds to optimization with repeated training, finding well generalizing parameters.
In the absence of pruning LRR follows a repeated cyclic training procedure, referred to as cyclic training for the rest of the paper. 
Such a training procedure also boosts the performance of a dense network above the peak obtained by LRR. 
While \cite{prune-regularize} has also realized that a similar training procedure like LRR, without pruning, could increase the performance of a dense network, they have focused on analyzing the regularization effect of pruning and found that pruning with LRR outperforms a dense network in the presence of label noise. 

However, we focus on the optimization benefits of the cyclic training procedure of LRR in the absence of label noise.
We find that dense networks usually outperform pruned networks with our improved cyclic training schedule, highlighting the dominant role cyclic training plays to achieve state-of-the-art performance with LRR. 
The central insight of our work is, however, that cyclic training substantially boosts the performance of pruning at initialization methods like SNIP \citep{snip} and Synflow \citep{synflow} as well as random masks \citep{liu2021unreasonable,er-paper} (see Figures ~\ref{fig:er-cyclic} and \ref{fig:cyclic-imagenet}). 
It is more effective in doing so than simply training the sparse masks for longer. 
Even potential regularization effects of sparsity that mitigate label noise can be realized on sparse masks with cyclic training.
These improved PaI masks not only consistently outperform or match LRR in the low sparsity regime, they also achieve state-of-the-art PaI performance in the high sparsity regime in spite of relying on fewer training cycles than LRR.
While cyclic PaI can still not compete with LRR at high sparsities, we  set out to understand its limiting factors and exploit its merits to enable sparse training even at high sparsity.

In this process, we challenge the assumption that LRR primarily excels at mask learning, as it can accurately measure the importance of trained parameters. 
Strikingly, we find that cyclically training a supposedly superior sparse LRR mask with a random initialization does not surpass a cyclically trained random mask (or other PaI masks).
As we find, it can still obtain LRR performance (with cyclic training) but only by relying on a parameter initialization that is sufficiently coupled to the mask identification process.
Conceptually, this is in line with insights into the lottery ticket hypothesis that suggest, iterative pruning also serves the purpose to identify an initialization that contains information about the task \cite{paul2023unmasking} or at least parameter signs that support retraining \cite{lrr-signs}. 

This suggests that the primary information missing in PaI is the right coupling between mask and parameter initialization.
To improve this coupling, we propose SCULPT-ing (\textbf{S}parse \textbf{C}yclic \textbf{U}ti\textbf{L}ization of \textbf{P}runing and \textbf{T}raining), as illlustrated in Figure ~\ref{fig:cyclic-train}.
It starts with a) cyclic training of a (potentially random) sparse mask, which b) is pruned in a single step and c) retrained with a single training cycle.  
This way, SCULPT-ing transfers the main benefits of iterative pruning, i.e., cyclic training and parameter-mask coupling to sparse training, while requiring fewer computational and memory resources at high sparsity.

Our main contributions are as follows:
\begin{itemize}
    \item We propose repeated cyclic training as an optimization procedure for sparsely initialized neural networks (including random ones), achieving state-of-the-art pruning at initialization performance. 
    \item Investigating the benefits over just training longer, we find that repeated cyclic training achieves lower training loss and better generalization performance by jumping between local optima and potentially finding a flatter loss landscape. 
    Cyclic training is more effective at finding effective parameter signs for sparse networks.
    \item At high sparsity, we highlight the importance of an appropriate coupling between parameter initialization 
    and the sparse mask to obtain state-of-the-art performance. 
    In the absence of coupling, we find that the mask learnt by iterative methods induces no benefits over a random mask.
    \item We propose SCULPT-ing to reach a similar performance as LRR but at reduced computational and memory costs by combining sparse cyclic training with one-shot pruning.
\end{itemize}


\section{Background and related work}

\textbf{Iterative pruning and lottery tickets.}
Iterative pruning methods entail an iterative training and pruning procedure to sparsify neural networks by removing parameters based on an importance measure, usually parameter magnitude.
\citep{han2015learning} empirically showed the success of these methods on CNNs.
\citep{frankle2019lottery} introduced the Lottery Ticket Hypothesis (LTH) and utilized Iterative Magnitude Pruning (IMP) to find sparse, trainable subnetworks of dense randomly initialized source networks, i.e., lottery tickets, that can be trained from scratch to achieve a similar performance as training the dense source network.
Although \citep{frankle2019lottery} show the existence of lottery tickets (LTs), they are only able to find them retrospectively by repeating the following steps: a) training a (dense) network, pruning usually 20\% of the parameters based on lowest magnitude, c) rewinding the remaining parameters to their initial value.
As this approach is less successful on more complex tasks and architectures, \citep{rewindVsFinetune} proposed Weight Rewinding (WR) and Learning Rate Rewinding (LRR), which obtain state-of-the-art performance for sparse networks across datasets with iterative pruning.
While IMP rewinds to initial weights, WR rewinds to a point obtained after a few training steps, and LRR continues training from the learnt weights of the previous iteration and thus never rewinds the learned neural network parameters.
This allows LRR to consistently outperform WR and IMP \citep{rewindVsFinetune, lrr-signs}, yet, we find that repeated cyclic retraining of the WR network is able to fill the gap between WR and LRR. 

\textbf{Task specificity of LT initialization.}
While the original LTH has given great hope that training sparse neural networks from scratch might be feasible, it has become evident that the mask of sparse LTs \cite{oneforall,universallanguage,uniExist} as well as the identified initial parameters contain task specific information \cite{paul2023unmasking} that is obtained only by training the dense overparameterized network and it is unclear how to identify mask and initialization otherwise.
Theoretical LT existence proofs \cite{malach2020proving,pensia2020optimal,orseau2020logarithmic,plant,depthexist,convexist,cnnexist,er-paper,ferbach2023a} even suggest that just pruning the random source network can perfectly couple the mask and its initial parameters so that no further training is required.
Full task specific information can even be contained in a subset of the parameters \cite{BNsufficient,fcBNexpressive,BNtheoryCNN}. 
But is a similar coupling between initialization and mask only attainable by iterative pruning?
We propose SCULPT-ing as an alternative.


\textbf{Benefits of iterative pruning.}
The main motivation of pruning is usually the reduction of computational resources, yet, it has also been found to lead to   improved generalization \citep{lecun1990optimal,hassibi1993optimal,frankle2019lottery,prune-regularize} at an optimal sparsity due to repeated training cycles and due to a regularization effect in the presence of label noise. 
We leverage an improved cyclic training procedure to enable sparse training from scratch.
Yet, the success of iterative pruning schemes has been attributed to their ability to transfer crucial information about the loss landscape between consecutive pruning iterations, as they are linearly mode connected \citep{paul2023unmasking,dugradient,frankle2021review}, to find a performant sparse mask and initialization pair.
While full training is not necessary to find a good mask \citep{earlybird},  
training with initial overparameterization in early pruning cycles has been conjectured to improve the mask identification and enable meaningful parameter sign flips during learning \citep{zhou2019deconstructing,lrr-signs}.
It is an open question whether PaI could enjoy similar advantages.
We find that cyclic training significantly increases the number of parameter sign flips from initialization. 

\textbf{Pruning at initialization (PaI).}
PaI methods aim to identify a sparse mask at initialization and enable sparse training from scratch.
They use an importance measure like connection sensitivity (SNIP) \citep{snip}, gradient signal preservation (GraSP)\citep{grasp} or criteria that maximize the number of paths while ensuring sufficient widths \citep{pham-paths, patil2021phew, synflow} to prune weights.
\citep{liu2021unreasonable, er-paper} also showed that random pruning is a simple and effective pruning at initialization method which was also earlier verified in sanity checks of mask learning \citep{sanity,sanity2}.
We boost their performance significantly with cyclic training.

\textbf{Training schedules.}
LRR relies on a repeated cyclical learning rate schedule that improves performance at certain sparsities as a consequence of repeated training cycles. 
Such cyclic training can also improve generalization of dense networks as observed by \citep{prune-regularize} and confirmed in Figure ~\ref{fig:lrr-peak}.
Its general benefits for dense training have been conjectured to be that cyclic training schedules can induce different benefits, including  \citep{smith2017cyclical}. 
Is this different from simply training longer?
Recent work by \citep{kuznedelev2023accurate} also suggests that sparse networks are under-trained and proposes training them for increased epochs with a linearly decaying learning rate. 
However, they attribute the success of increased training to better mask exploration for methods like RiGL \citep{evci2019rigging} and AC/DC \citep{peste2021acdc}, which dynamically update the mask during training, and not for a fixed PaI mask which is the focus of our work.
Interestingly, we find that simply training longer is not as effective as cyclic training for PaI.

\section{Repeated cyclic sparse training}
\begin{figure}[h!]
 \centering
  \includegraphics[width=
\textwidth]{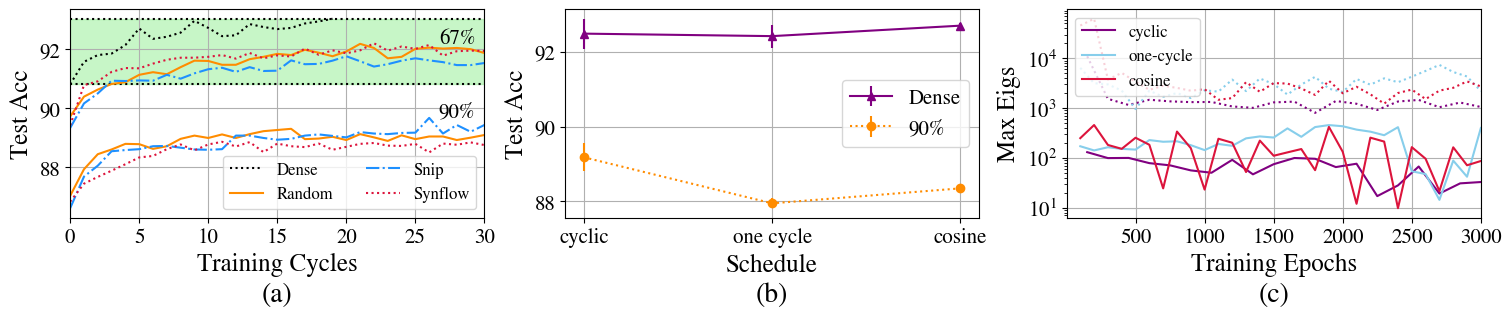}
   \caption{(a) Improved generalization by cyclic training of a sparse mask (with sparsity $67\%$ and $90\%$) and a dense network. 
   (b) Cyclic training improves over training with a one-cycle learning rate schedule and for a dense network and a random sparse network with $90\%$ sparsity. (c) Maximum eigenvalues of the Hessian of the loss function for a dense network (solid lines) and  random sparse network with $90\%$ sparsity (dotted lines). Results are reported for CIFAR10.}
  \label{fig:er-cyclic}
\end{figure}

Iterative pruning methods like LRR enjoy the additional benefit of improved generalization performance in comparison with a dense network as shown in Figure \ref{fig:lrr-peak} \cite{frankle2019lottery,rewindVsFinetune}. 

\textbf{Cyclic training improves generalization.} 
\citep{prune-regularize} conjectured that parts of this improvement could be attributed to the LRR training schedule but focused their analysis on the additionally induced regularization effect of pruning. 
To verify that repeated cyclic training benefits generalization, we first train a dense network, without pruning, for the same number of cycles as LRR by repeating the training schedule in each cycle.
The black dotted line in Figure \ref{fig:lrr-peak} denotes the improvement in generalization of the dense network with cyclic training over standard training.
The dense network sees an increase in performance in the first few cycles, before it plateaus, indicating that only a few additional training cycles are needed to improve the optimization. 

\textbf{Insights into mechanisms of cyclic training.}
Complementary to \citep{prune-regularize}, we argue that cyclic training has a strong influence on LRR and also boosts dense training, as suggested by \cite{smith2017cyclical}. 
It also seems to define a generally advantageous learning rate schedule that truly shows its merits in the context of sparse training, which we aim to exploit here. 
We dedicate this section to investigate the potential mechanisms that could explain its superior performance. 

Concretely, we discuss four different but related hypotheses.
1) Training for more epochs is simply better in optimizing the training and test loss. In particular, in the high sparsity regime, \cite{kuznedelev2023accurate} have encountered that networks tend to be under-trained ,in the context of a different pruning method. 
2) The regular increase of the learning rate allows cyclic training to effectively jump between local optima in the loss landscape and find flatter optima that have been associated with better generalization \cite{hochreiter1997flat}.
3) Cyclic training is more flexible in flipping and learning meaningful weight signs, a task at which LRR was conjectured to excel \cite{lrr-signs}.
4) Cyclic training improves the conditioning of the learning task. 
As it turns out, all four provide a partial explanation, but 2) seems to be the most distinguishing factor of cyclic training, as we show in the following.


\textbf{Training longer.}
The overall training procedure of LRR takes more training epochs than usual, also because the training cycles have to compensate for pruning operations. Could simply training for longer already improve the generalization performance?
To test this hypothesis, we compare cyclic training with two other learning rate schedules, a common one cycle \citep{adaptive-lr} and cosine schedule, which we extend over the same number of training epochs, as visualized in Figure~\ref{fig:lr_schedules_compare}.
Note that the cosine schedule also consists of multiple cycles and thus shares the basic features of cyclic training, yet, the cycle itself is different.  
Interestingly, according to Figure~\ref{fig:er-cyclic}~(b), cyclic and cosine schedules are similarly effective in training a dense network and outperform one cycle training, suggesting that the exact LRR schedule might be less special than previously assumed \cite{prune-regularize}.
Yet, the cyclic training achieves best generalization on a $90\%$ sparse mask, promising higher gains in the context of sparse training.
Figure~\ref{fig:er-cyclic}~(a) confirms a considerable performance boost over standard training also for other PaI masks resulting from cyclic training. 



\begin{figure}[h!]
    \centering
    \includegraphics[width=\textwidth]{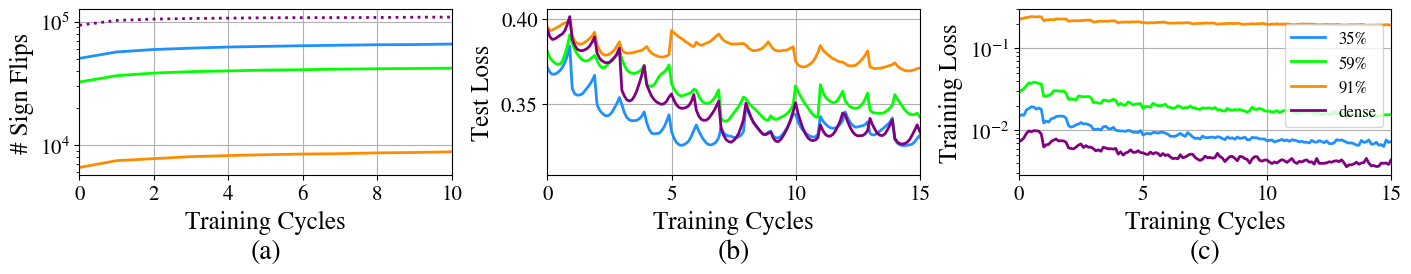}
    \caption{Cyclic training of a random sparse mask on CIFAR10 with a ResNet20 at different sparsities. (a) Number of sign flips during training. (b) Linear mode connectivity of the test loss and (c) train loss after consecutive training cycles. 
    }
    \label{fig:signs}
\end{figure}

\textbf{Flatness and conditioning.}
Increasing the learning rate during training could help escape local minima, which we confirm by a linear mode connectivity analysis.
Interpolating the networks after every cycle for a dense and random sparse network, we observe that consecutive cycles often have linearly connected test loss, however, the train loss is separated by an error barrier.
This suggests that cyclic training is able to escape local optima allowing an improved exploration of the loss landscape.
An increased number of cycles also reduces the error barrier of the training loss between consecutive cycles, pointing at a shift towards flatter optima (see Figure \ref{fig:signs}(b), (c)).
In contrast, the train loss plateaus earlier for the cosine schedule, while one-cycle training jumps fewer optima (see Figure \ref{fig:lmc-cosine}).
%

By approximating the largest eigenvalue of the hessian of the loss function as a proxy for flatness, we further confirm in Figure \ref{fig:er-cyclic}(e) that cyclic training ends in a flatter neighborhood, which is known to correlate with improved generalization \citep{hochreiter1997flat,keskar2016large, dziugaite2017computing}. 
In addition, the decrease of the maximum eigenvalue in the course of training suggests that also the learning task becomes better conditioned over time and that easier to solve subsequent training cycles.

\textbf{Sign flips.}
\citep{lrr-signs, zhou2019deconstructing} suggest that LRR is capable of improved parameter optimization because of its flexibility to learn task-relevant parameter signs.
Moreover, they find that the majority of the sign flips occur in the early iterations of LRR.
\citep{lrr-signs} conjectures that the overparameterization of networks in early iterations enables the sign flips, but is cyclic training the real cause? 
We plot the number of parameter sign flips from initialization and find concurring results with cyclic training for sparse networks.
Cyclic training increases the number of sign flips compared to standard training, but the majority of sign flips occur in the earlier cycles (see Figure \ref{fig:signs}(a)), potentially allowing them to generalize better.
We also find that cyclic training schedules are better at recovering correct signs as compared to one-cycle (see Appendix \ref{app:signs}).

\begin{table}[h!]
\caption{Comparison of the number of training cycles required for LRR to reach a target sparsity, performance of cyclic training of a random mask with the same number of cycles as LRR and cyclic training of a random mask with increased cycles till performance peaks, on CIFAR10.}
\label{tab:peak}
    \centering
    \begin{tabular}{c | ccc | cc}
        \hline
        Sparsity & \# cycles (LRR) & Acc (LRR) & Acc (cyclic) &  \# cycles to peak &  Acc (cyclic) \\
        \hline
        $59\%$ & $5$  & $92.00$ & $92.20$ & $12$ & $92.29$ \\
        $79\%$ & $8$  & $91.71$ & $90.51$ & $14$ & $90.99$ \\
        $95\%$ & $15$ & $89.06$ & $85.67$ & $14$ & $85.86$ \\
        \hline
    \end{tabular}
\end{table}

\begin{figure}[h!]
    \centering
    \includegraphics[width=0.9\textwidth]{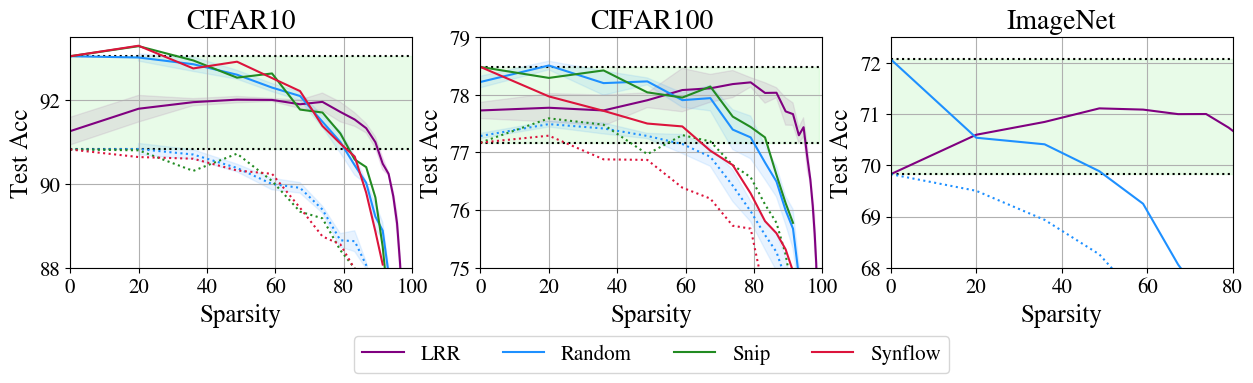}
    \caption{Cyclic training boosts performance of any sparse mask including a random one and even outperforms LRR at low sparsity. Shaded region highlights the gain in performance of a dense network by cyclic training for reference. Solid lines denote results with cyclic training and dotted lines show standard training for PaI methods.}
    \label{fig:cyclic-imagenet}
\end{figure}

\textbf{Boosting PaI performance with cyclic training.} 
Having established the benefits of cyclic training, we propose to exploit it for training sparse masks identified at initialization with PaI, which we term cyclic PaI. 
Figures \ref{fig:er-cyclic} and \ref{fig:cyclic-imagenet} show that, similar to a dense network, cyclic training also improves the generalization of a sparse network, but the boost is more significant at higher sparsity.
Unlike in LRR where the number of training cycles depends on the final sparsity, we can adjust the number of training cycles to trade-off performance with computational savings. Table~\ref{tab:peak} suggests that LRR potentially under-trains its masks at lower sparsities. 
This enables cyclic PaI to outperform LRR at low sparsity (see also Figure~\ref{fig:coupling}). 
At higher sparsity, cyclic PaI needs potentially fewer training cycles than LRR.
Yet, it can only match the performance of LRR on ImageNet at $20\%$ sparsity beyond which the effect of pruning becomes more important.



\textbf{Regularization effect of cyclic PaI.}
Complementing the finding by \citep{prune-regularize} that LRR pruning increases robustness to label noise, we find that cyclic PaI can realize similar benefits but with initial sparsity according to Figure \ref{fig:sculpt-criterion-noise}(b)), where a sparse random network generalizes better than a dense one.


\textbf{Relevance of the mask.}
We observe that different choices of sparse masks using criteria like SNIP or Synflow seem equivalent with cyclic training. 
Similar conclusions were obtained also in the absence of cyclic training  \citep{liu2021unreasonable,sanity}.
Yet, cyclic PaI is unable to compete with LRR in the high sparsity region.
This gap is most pronounced on ImageNet where, although cyclic training improves a random mask considerably, it still falls short compared to LRR.
As the optimization procedure for both LRR and a random mask is now similar, the only difference between them seems to be the sparse mask.
However, can we really attribute the gap between LRR and cyclic PaI to task-specific mask learning?
As we see in the next section, this conclusion would overlook the central role of the parameter initialization. 



\textbf{Conclusion.} From this section we conclude that cyclic training can significantly boost PaI methods and even outperform LRR in low sparsity regions, which provides a proof of principle that a strong optimization scheme can make sparse training competitive. 
The following section seeks to uncover why LRR still performs better in the high sparsity regime. 

\section{Does the mask matter?}
\label{sec:coupling}

\begin{figure}[h!]
    \centering
    \includegraphics[width=0.9\textwidth]{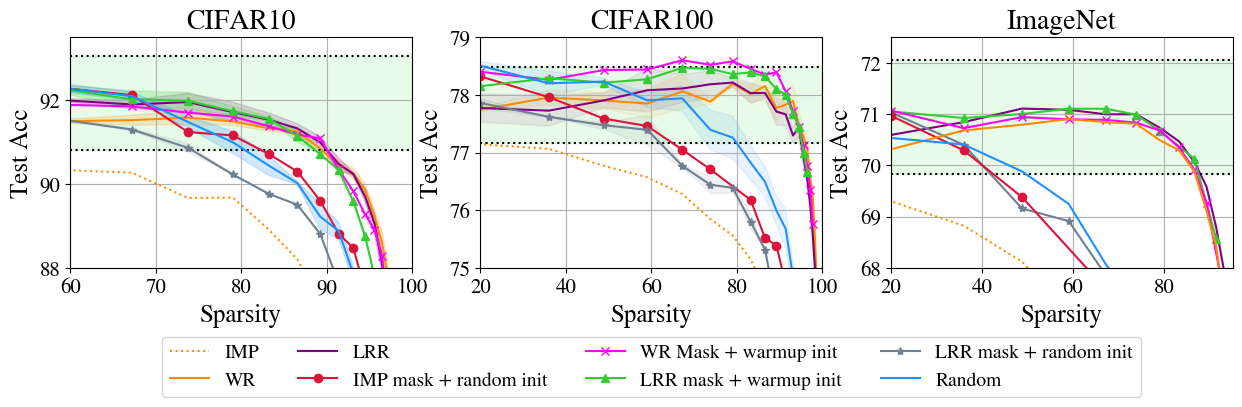}
    \caption{Comparing cyclic training with different combinations of sparse masks and parameter initializations to iterative pruning methods LRR, WR and IMP.}
    \label{fig:coupling}
\end{figure}
Having established that the learning rate schedule of LRR drives most but not all of its performance, we are left to wonder what constitutes its strength in the high sparsity regime. 
The obvious difference between cyclic PaI and LRR are the masks that are optimized. 
As illustrated in Figure \ref{fig:cyclic-train}, iterative pruning gradually removes the parameters with smallest magnitude in every iteration, thus learning a potentially task specific sparse mask.
In contrast, PaI methods identify a sparse mask in a single pruning step at initialization, based on potentially less accurate information.

\textbf{LRR learns more than mask structure.}
Investigating the sparse mask learnt by LRR, we initialize it with a new random initialization, followed by cyclic training.
To our surprise, we observe that the mask identified by LRR with a random initialization is no better than a random mask after cyclic training, as shown in Figure \ref{fig:coupling} (LRR mask + random init).
However, if we initialize the learnt LRR mask with the parameters of a dense network that was trained for a few steps, like in WR, and then perform cyclic training on this combination (LRR mask + warmup init), we are able to recover baseline LRR performance even at high sparsity. 
This suggests that along with improved optimization via cyclic training, it is crucial to have an appropriate initialization for the sparse mask to improve performance at high sparsity.
It also implies that the mask structure learnt by LRR might not be special on its own, but is in combination with the parameter initialization.

\textbf{Coupling of parameter initialization and mask.}
In order to identify the combinations of parameter initialization and mask that can match LRR at high sparsity, we also look at the masks and initializations of the other iterative pruning methods WR and IMP and optimize each of these mask parameter pairs with cyclic training.
The key difference between WR and IMP is that after each iteration, IMP rewinds its parameters to their initial values while WR rewinds to values trained for a few steps (denoted by warmup init).
Cyclic training of an IMP mask combined with its random initialization (IMP mask + random init) is able to improve over standard IMP, however is only at par with cyclic training of a random mask. 
Whereas, cyclic training of a WR mask combined with its warmup initialization (WR mask + warmup init) is able to match the performance of LRR, similar to LRR mask + LRR init.
These results, shown in Figure \ref{fig:coupling}, also confirm that when the mask and parameters are coupled, for example as in case of a warmed up initialization and an iteratively learnt mask, they can match the performance of LRR with cyclic training.
This insight is particularly interesting, as it suggests that lottery tickets might also exist that can achieve LRR performance.

\begin{figure}[h!]
    \centering
    \includegraphics[width=\textwidth]{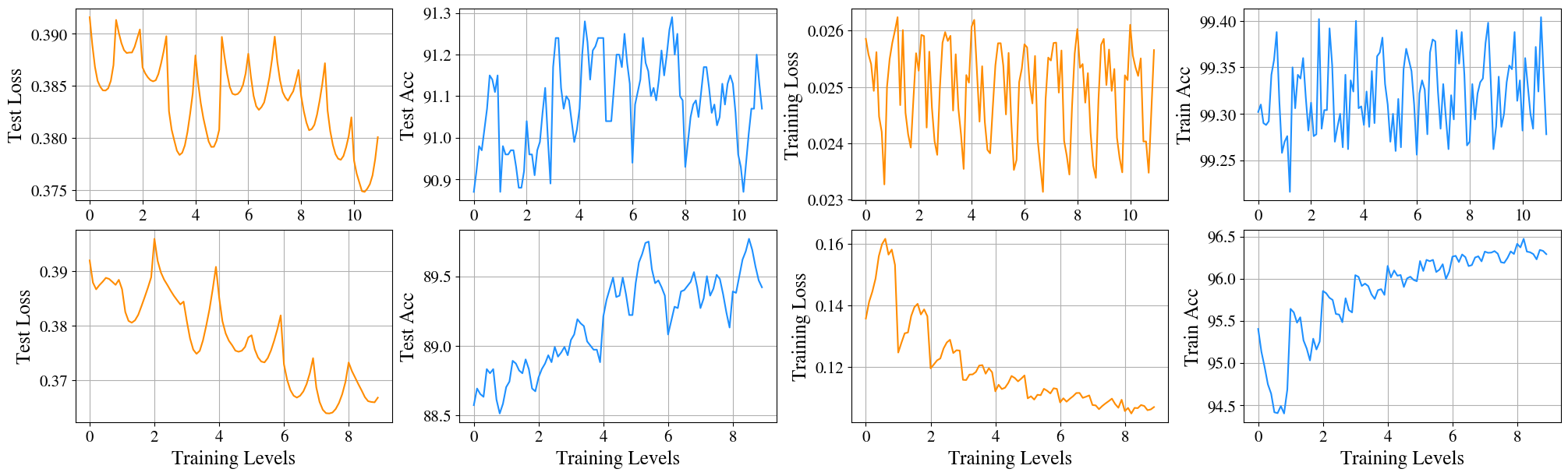}
    \caption{Linear mode connectivity of consecutive training cycles for LRR mask + warmup init (top) and LRR mask + random init (bottom) on CIFAR10.}
    \label{fig:lmc-coupling}
\end{figure}

However, is cyclic training really required to achieve this high performance?
A linear mode connectivity analysis in Figure \ref{fig:lmc-coupling} further sheds light on the coupling phenomenon.
In the case of an LRR mask + random init, consecutive cycles have linearly connected test loss while the training loss has error barriers between cycles.
However, for LRR mask + warmup init, we see that consecutive cycles are mostly in the same loss basin, at least at later stages and enable matching the performance of LRR (see also Figure \ref{fig:c100-lrr-random}).
An initialization that is coupled to the mask and is task specific, starts in the final loss basin or close to it.
LRR and WR are known to follow a similarly linearly mode connected optimization trajectory \citep{paul2023unmasking}, while IMP does not enjoy the same benefit, as it always restarts from a random initialization, and struggles to keep up with the performance of LRR and WR, which is in line with our coupling analysis (see also Figure \ref{fig:lmc-coupling-iterative}).


\textbf{Conclusion.} Cyclic training alone is not sufficient to succeed at high sparsity, but requires an initialization that is well coupled to a mask.
Our analysis is inconclusive whether LRR masks alone are better aligned with a learning task than PaI masks and poses the potential universality of lottery tickets in the high sparsity regime as an open question \cite{oneforall,universallanguage,uniExist}. 





    

\section{SCULPT-ing}

Our empirical investigations so far have highlighted the potential of cyclic PaI to act as  sparse training paradigm, yet, it lacks the right parameter initialization for a given mask and task to compete in the high sparsity regime.  
Only LRR and to some extent WR have been able to realize the benefits of both the right  initialization-mask coupling and cyclic training, as they consistently find highly performant sparse networks.
However, both LRR and WR are computationally demanding and memory intensive as they start from a dense network.
To enhance sparse training and address the coupling issue, we propose SCULPT-ing, which can achieve a similar performance as LRR and WR while starting sparse network and requiring fewer training cycles at high sparsity.
Our experiments verify that SCULPT-ing is often able to bridge the gap between cyclic PaI and LRR at high sparsity.

\textbf{SCULPT-ing.} 
(a) Find a sparse mask at initialization with PaI method of choice. 
(Our experiments focus on a random mask.)
(b) Train with cyclic training to reach peak performance or for the same number of epochs that LRR would take to reach the initial sparsity.
(c) Sparsify further by a single step of magnitude pruning to obtain the final sparsity.  
(d) Retrain with only one training cycle.

The magnitude based pruning step in (c) serves the purpose to couple the learnt parameters to the task and the final sparse mask.

\begin{figure}[h!]
    \centering
    \includegraphics[width=0.9\textwidth]{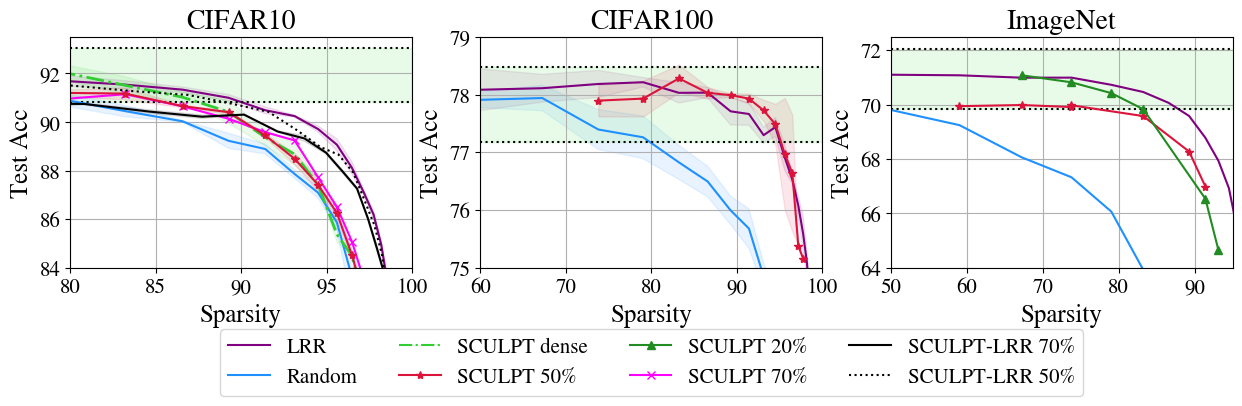}
    \caption{SCULPT-ing results starting from a random sparse mask of different sparsities.}
    \label{fig:sculpt}
\end{figure}

%

\textbf{Experimental results. }
SCULPT-ing results are shown in Figure \ref{fig:sculpt}.
On CIFAR100, SCULPT matches LRR performance, while starting with a $50\%$ sparse mask.
On ImageNet, SCULPT can match LRR starting from a $20\%$ sparse mask, while still being competitive if it starts from a $50\%$ sparsity.
On CIFAR10, we can start SCULPT-ing as sparse as $70\%$ and outperform cyclic training of a random mask.
Yet, it is unable to match the performance of LRR, but this gap can be closed by combining SCULPT-ing with LRR, i.e., cyclic training of a sparse network and performing LRR subsequently.
Interestingly, LRR is less amenable to early sparsification than other pruning approaches \cite{er-paper}, which aligns with the conjectured conjectured benefits of early overparameterization \cite{lrr-signs}.

\textbf{Training time.}
SCULPT-ing allows sparse networks trained from scratch to compete with and match the performance of LRR and offers two benefits.
First, always training a sparse network allows a smaller memory footprint in contrast to LRR which starts from a dense network.
Second, the number of training cycles for LRR depends on the final sparsity of the network, i.e., a higher sparsity requires more cycles as every cycle prunes only $20\%$ of nonzero parameters.
SCULPT-ing however uses a flexible number of training cycles for any sparsity and can thus reduce total training cycles at high sparsity.
We choose the number of training cycles in SCULPT-ing to maximally boost performance of the sparse mask followed by one additional cycle of retraining after pruning. 
The initial number of training cycles can be traded-off for a smaller boost to further reduce the training time.

\textbf{Computational savings.}
In experiments, we train SCULPT with cyclic training for $14$ cycles, i.e., $14 \times 150 = 2100$ epochs for CIFAR10 and CIFAR100 and $6$ cycles, i.e., $6 \times 90 = 540$ epochs for ImageNet, for each sparsity.
In comparison, LRR reaches a sparsity of $100(1 - 0.8^{14})= 95\%$ for CIFAR10 and $100(1-0.8^{6}) = 74 \%$ for ImageNet in the same number of cycles, assuming $20\%$ parameters are pruned in each cycle.
Hence, SCULPT-ing at sparsity greater than $95\%$ for CIFAR10 and CIFAR100 and $74\%$ for ImageNet will take fewer cycles than LRR.
At low sparsities, cyclic training is able to boost performance over LRR while maintaining a smaller memory footprint at the cost of increased training cycles, which can also be exploited by SCULPT-ing if desired.

\begin{figure}[h!]
\centering
 \includegraphics[width= 0.8
\textwidth]{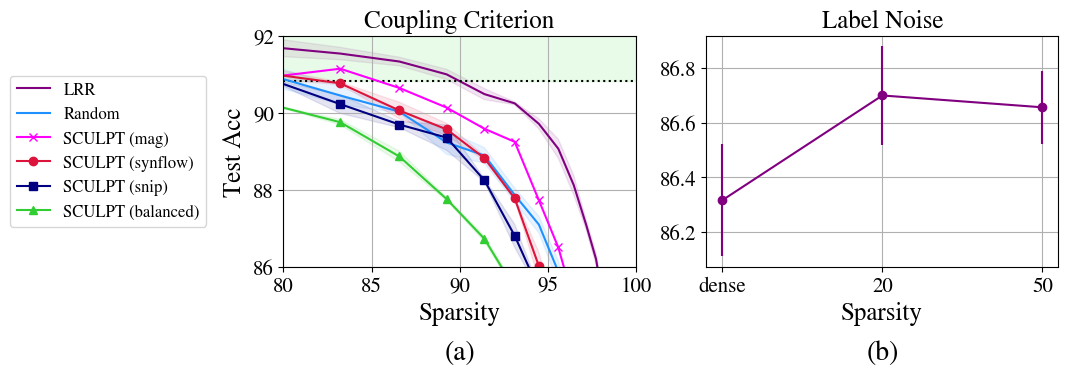}
  \caption{(a) Coupling with different pruning criteria with SCULPT-ing on CIFAR10 starting from a $70\%$ sparse random mask. (b) Regularization effect of a random sparse mask with cyclic training on CIFAR10 with $15\%$ label noise.}
  \label{fig:sculpt-criterion-noise}
 \end{figure}

\textbf{Magnitude pruning enables coupling.}
Figure \ref{fig:sculpt-criterion-noise}~(a) investigates alternatives to magnitude based pruning in the one-shot pruning step of SCULPT-ing.
Interestingly, magnitude seems to be best suitable for realizing a good coupling between mask and its parameters. 
This might be explained by the finding that magnitude based pruning minimally changes the neural network function \citep{cosine-magprune}. 


\textbf{Experimental Setup.}
\label{sec:expt}
All empirical investigations are performed on image classification tasks, to validate our insights.
We train a ResNet20 network for the CIFAR10 \cite{cifar10} dataset and use a ResNet18 \cite{resnet} for CIFAR100 and ImageNet \cite{deng2009imagenet} datasets.
Our networks were trained on NVIDIA A100 GPUs.
All experimental details are provided in Appendix \ref{app:setup}. 
Accuracy curves in Figure \ref{fig:cyclic-imagenet}, \ref{fig:coupling},\ref{fig:sculpt} and \ref{fig:sculpt-criterion-noise} are reported with respect to sparsity i.e. the fraction of zeroed out (pruned) parameters in the network.
Sparsity is also given by 1 - density, where density is the fraction of non-zero parameters.

\section{Discussion}
\label{sec:discuss}

We have conducted a rigorous empirical investigation into the inner mechanisms of state-of-the-art iterative pruning methods Learning Rate Rewinding (LRR) and Weight Rewinding (WR)
While their superior performance has largely been attributed to improved mask identification and an implicit sparsity regularization, we have challenged this belief and  presented evidence for the insight that their repeated cyclic training schedule enables improved optimization.

To transfer its merits to sparse training, we have proposed to combine cyclic training with pruning at initialization (PaI), which can outperform even LRR at lower sparsity. 
The performance boost is particularly striking, as \cite{lrr-signs} conjectured that mainly early overparameterization supports LRR in learning sparse, highly performant models.
As it turns out, a relevant share of its performance and ability to flexibly switch signs is induced by its cyclic training procedure.

Yet, cyclic PaI also faces limits in the high sparsity regime, where we find no significant performance differences between masks, including a mask that has been identified by LRR and can, in principle, achieve a higher performance. 
This finding identifies a remaining challenge of cyclic PaI, i.e., deriving a parameter initialization that is sufficiently coupled to the mask and learning task so that cyclic training can effectively learn in the high sparsity regime. 

To improve this coupling in the context of sparse training, we have proposed SCULPT-ing, which performs cyclic training of a sparse mask followed by a single magnitude based pruning step to induce the desired coupling.
SCULPT-ing bridges the gap between sparse training and iterative pruning to save computations in comparison with LRR and improve the performance of cyclic PaI.

While SCULPT-ing can solve a trade-off between computational and performance considerations by adapting its number of training cycles, efficient sparse training remains a challenge that asks for further insights into improved mask identification and effective parameter optimization.

\section{Acknowledgements}

We gratefully acknowledge funding from the European Research Council (ERC) under the Horizon Europe Framework Programme (HORIZON) for proposal number 101116395 SPARSE-ML.

\bibliography{example_paper}
\bibliographystyle{icml2024}


\appendix
\newpage
\section{Appendix}

\subsection{Improved sign recovery by cyclic training.}
\label{app:signs}
\begin{figure}[h!]
    \centering
    \includegraphics[width=0.6\textwidth]{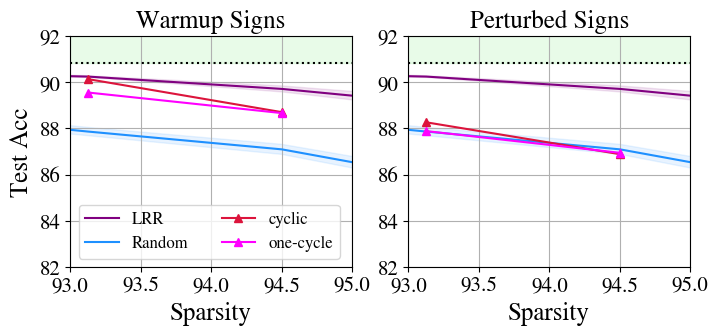}
    \caption{Ability to recover signs for cyclic training in comparison to one-cycle. Results for training a learnt LRR mask with the signs of warmup initialization and random magnitudes (left) and the same with the $20\%$ of the signs also randomly perturbed (right).}
    \label{fig:perturb}
\end{figure}

Given the importance of sign flips for sparse training \citep{lrr-signs}, we investigate if cyclic training is better at recovering correct signs as compared to one-cycle training for the same number of epochs.
We use the coupling experimental setup from Section \ref{sec:coupling} and train a learnt LRR mask with the signs of the warmup initialization and randomized magnitude with both cyclic and one-cycle training, as reported in Figure \ref{fig:perturb}(b).  
We find that, with the warmup signs, cyclic training is exactly able to recover LRR performance while one-cycle is worse, while at higher sparsity both cyclic and one-cycle perform identically.
Similarly, perturbing $20\%$ of the initial signs in the same also shows that cyclic training can recover better at lower sparsity but is identical to one-cycle at high sparsity.

To further examine the ability of sign recovery, we find that the signs learnt by cyclic training have a $95.37\%$ overlap with the signs learnt by LRR whereas one-cycle has an overlap of $93.67\%$.
A higher overlap with cyclic training suggests that it is better at being able to recover the signs given the signs at warmup.

However, it is also important to note that LRR is also trained cyclicly, which might be the reason why cyclic training of the warmup signs is able to recover the same signs better.

We also find that for an LRR mask + warmup init at $93\%$ sparsity, if $20\%$ of the signs are randomly perturbed, cyclic training is able to recover to an accuracy of $88.71\%$ as compared to $88.12\%$ with one-cycle training for $2000$ epochs each. 

\subsection{Experimental Setup}
\label{app:setup}
The codebase for our experiments was written using PyTorch and torchvision and their relevant primitives for model-construction and data-related operations. In the context of ImageNet experiments we made use of FFCV \citep{leclerc2023ffcv} for fast dataloading. All models used to report the numbers in in the experiments were trained on a single NVIDIA A100 GPU. We provide all code for our experiments.

We report mean and $95\%$ confidence intervals over $3$ seeds for each run in our experiments, except the coupling experiments on CIFAR100 reported in Figure \ref{fig:cyclic-imagenet} and all runs on ImageNet for which we report single runs. 
All experiments used the SGD optimizer with a weight decay of $1e-4$ and momentum $0.9$. The batch size was fixed to $512$ across all experiments and datasets.

\begin{figure}[h!]
    \centering
    \includegraphics[width=\textwidth]{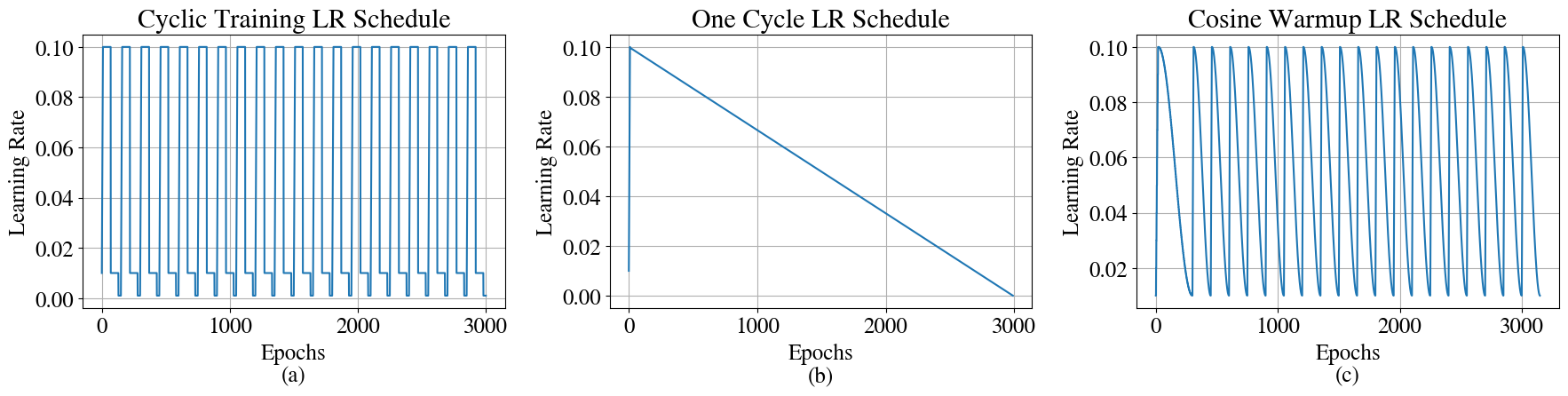}
    \caption{\textbf{Left:} The Step Warmup learning rate schedule for a single cycle, initially there is a linear warmup and subsequently there are two steps by a factor of $10$ \textbf{Middle:} Cyclic Training Learning rate schedule with multiple cycles the schedule in the left plot. \textbf{Right:} One Cycle learning rate schedule which uses a fixed cycle over 3000 epochs.}
    \label{fig:lr_schedules_compare}
\end{figure}

When using cyclic training, multiple cycles are used at each sparsity level.
Each cycle followed a learning rate schedule as shown in Figure \ref{fig:lr_schedules_compare}(a).

\begin{figure}[h!]
    \centering
    \includegraphics[width=0.6\textwidth]{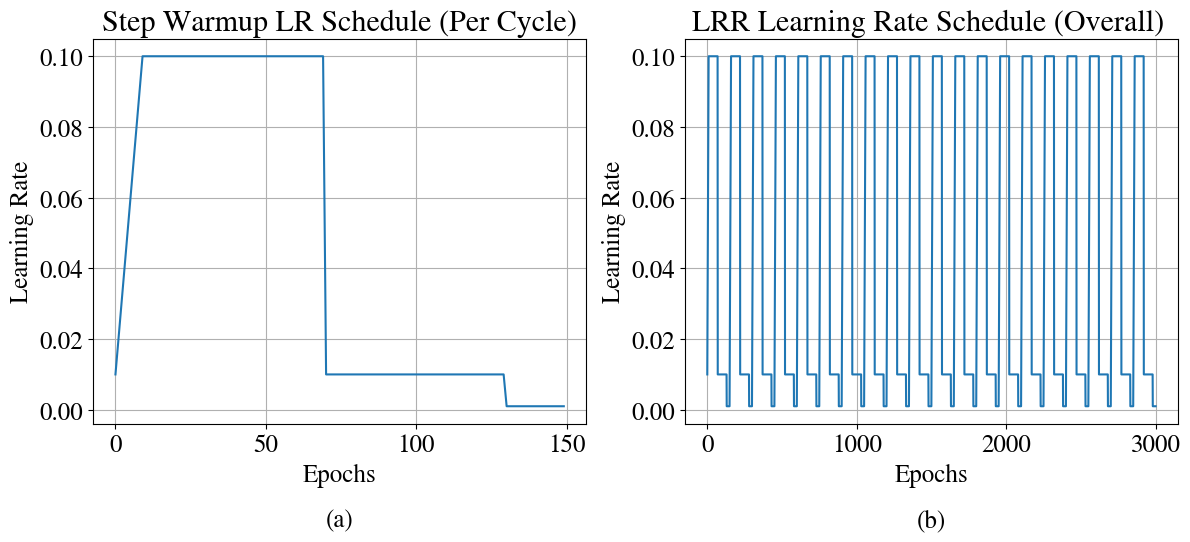}
    \caption{Learning rate schedule when performing standard LRR training across multiple levels. Here we train for one cycle per training level.}
    \label{fig:schedule_lrr}
\end{figure}

For CIFAR100 and CIFAR10 experiments, each individual training cycle used a multi-step warmup lr scheduler, which starts with a linear-warmup. Each individual cycle has a length of 150 epochs. Subsequent to the warm-up, from an initial learning rate of $0.1$, there is reduction by a factor of $10$ at epoch $70$ and $130$. 
For ImageNet, the cycle length was 90 epochs with a constant warmup for $10$ epochs followed by a step schedule at every $30$ epochs with a drop by a factor of $10$.

In \cref{fig:er-cyclic}, the max eigenvalues were computed using the PyHessian library \citep{pyhessian}.

\textbf{Label Noise:} The label noise experiments for CIFAR10 were carried out by randomly flipping 15\% using a random permutation of the labels to not impact the balance of labels across the train dataset. The test dataset remains uncorrupted.

\subsection{Iterative Magnitude Pruning (IMP)}
Iterative Magnitude Pruning was introduced by \citep{frankle2019lottery} paper. The pruning method can be described as follows:
\begin{itemize}
    \item Start with an initial dense network $f(x; \theta)$ where $\theta$ drawn from a distribution $D_{o}$. The objective is to find a mask $m$, to have a network $f(x;m \odot \theta)$ which is sparse.
    \item This model is then trained as usual, using an algorithm like stochastic gradient descent.
    \item The parameters of the trained network are then globally ranked according to their magnitude. Then $x\%$ of the lowest valued parameters are set to zero in the mask $m$ which has the exact same size as the network. Typically, $x = 20\%$.
    \item The parameters that have not been pruned (non-zero) after a pruning level are reset to the initial  random initialization $\theta_{o}$.
    \item This model is now trained again, repeating steps 2 - 4 until a target sparsity is reached. 
\end{itemize}

\subsection{Weight Rewinding (WR)}
One key challenge noticed with IMP in the lottery tickets paper was finding lottery tickets in deeper networks (VGG16 and ResNet). Lottery tickets were found at lower sparsities with use of a learning rate warmup, but there were none found at higher sparsities. So the authors of \citep{rewindVsFinetune} presented an alternate approach which worked much better. The tickets found are now called "matching tickets".

\begin{itemize}
    \item Start with an initial dense network $f(x; \theta)$ where $\theta$ drawn from a distribution $D_{o}$. The objective is to find a mask $m$, to have a network $f(x;m \odot \theta)$ which is sparse.
    \item The model parameters $\theta_{k}$ are saved at the $k^{th}$ epoch of dense training (usually after a warmup) that is now used as the rewound initialization. 
    \item This model is then trained as usual, using an algorithm like stochastic gradient descent.
    \item The parameters of the trained network are then globally ranked according to their magnitude. Then $x\%$ of the lowest valued parameters are set to zero in the mask $m$ which has the exact same size as the network. Typically, $x = 20\%$. This network can be represented by $f(x ; m \odot \theta)$
    \item The parameters that have not been pruned (non-zero) after a pruning level are are now "rewound" to their value in the weight parameters $\theta_{k}$.
    \item This model is now trained again, repeating steps 2 - 4 until a target sparsity is reached. 
\end{itemize}

\subsection{Learning Rate Rewinding (LRR)}
Learning rate rewinding introduced in \cite{rewindVsFinetune}, instead of resetting/rewinding to the relevant initialization as described above, allows the non-zero parameters to retain their learned values. Instead, LRR at every pruning level resets the learning rate schedule.

\begin{itemize}
    \item Start with an initial dense network $f(x; \theta)$ where $\theta$ drawn from a distribution $D_{o}$. The objective is to find a mask $m$, to have a network $f(x;m \odot \theta)$ which is sparse.
    \item The model parameters $\theta_{k}$ are saved at the $k^{th}$ epoch of dense training (usually after a warmup) that is now used as the rewound initialization. 
    \item This model is then trained as usual, using an algorithm like stochastic gradient descent.
    \item The parameters of the trained network are then globally ranked according to their magnitude. Then $x\%$ of the lowest valued parameters are set to zero in the mask $m$ which has the exact same size as the network. Typically, $x = 20\%$. This network can be represented by $f(x ; m \odot \theta)$
    \item This model is now trained again, retaining the learned values of the non-zero weights -- repeating steps 2 - 3 until a target sparsity is reached. 
\end{itemize}

\begin{table}[h!]
    \centering
    \begin{tabular}{| c | c | c | c |} 
 \hline
 Dataset &  CIFAR10 & CIFAR100 & ImageNet\\ [0.5ex] 
 \hline
 Model & ResNet20 & ResNet18 & ResNet18 \\
 \hline
 Epochs & 150 & 150 & 90\\ 
 \hline
 LR & 0.1 & 0.1 & 0.1\\
 \hline
 Scheduler & step-warmup & step-warmup & step-warmup \\
 \hline
 Batch Size & 512 & 512 & 512 \\
 \hline
 Optimizer & SGD & SGD & SGD \\
 \hline
 Weight Decay & 1e-4 & 1e-3 & 1e-4 \\
 \hline
 Momentum & 0.9 & 0.9 & 0.9 \\
 \hline
 Init  &Kaiming Normal & Kaiming Normal & Kaiming Normal \\ [1ex] 
 \hline
\end{tabular}
    \caption{Experimental Setup}
    \label{tab:expt-setup}
\end{table}

\subsection{Training iterations for cyclic training and LRR.}
Figure \ref{tab:computation} denotes the total number of training epochs required for LRR and for SCULPT for each sparsity.

\begin{table}[]
    \centering
    \begin{tabular}{c | c c c} 
 Dataset & LRR  & SCULPT \\
 \hline\hline
 CIFAR10 & $150 \times \text{\# iters}$ & $2000$ \\ 
 CIFAR100 & $150 \times \text{\# iters}$ & $2000$  \\
 ImageNet & $90 \times \text{\# iters}$ & $540$ \\
 \end{tabular}
    \caption{Number of training epochs required for LRR vs SCULPT-ing.}
    \label{tab:computation}
\end{table}

\subsection{ERK vs Balanced sparsity ratios}
\begin{figure}[h!]
    \centering
    \includegraphics[width=0.35\textwidth]{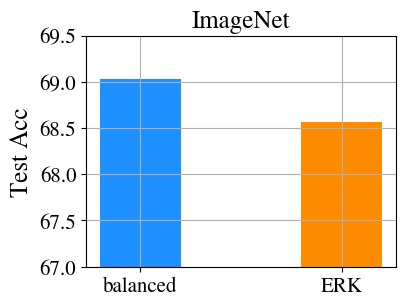}
    \caption{Random masks with different layerwise sparsity ratios on a ResNet18 trained on ImageNet.}
    \label{fig:erk-imagenet}
\end{figure}

We find that balanced layerwise sparsity ratios \cite{er-paper} find better random sparse masks than ERK sparsity ratios \cite{mocanu2018set} as shown in Figure \ref{fig:erk-imagenet}.

\subsection{Linear mode connectivity for cyclic training.}

We provide additional linear mode connectivity plots in support of our claims on the benefits of cyclic training and the importance of coupling.

\begin{figure}[h!]
    \centering
    \includegraphics[width=\textwidth]{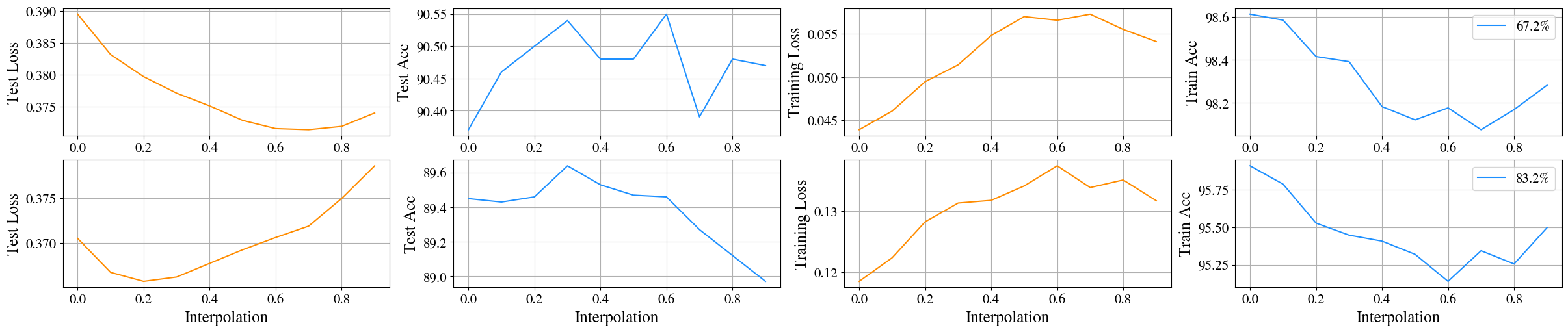}
    \caption{Linear mode connectivity of random networks after standard training i.e. one cycle of training. Each row corresponds to a sparsity.}
    \label{fig:lmc-1}
\end{figure}

\begin{figure}[h!]
    \centering
    \includegraphics[width=\textwidth]{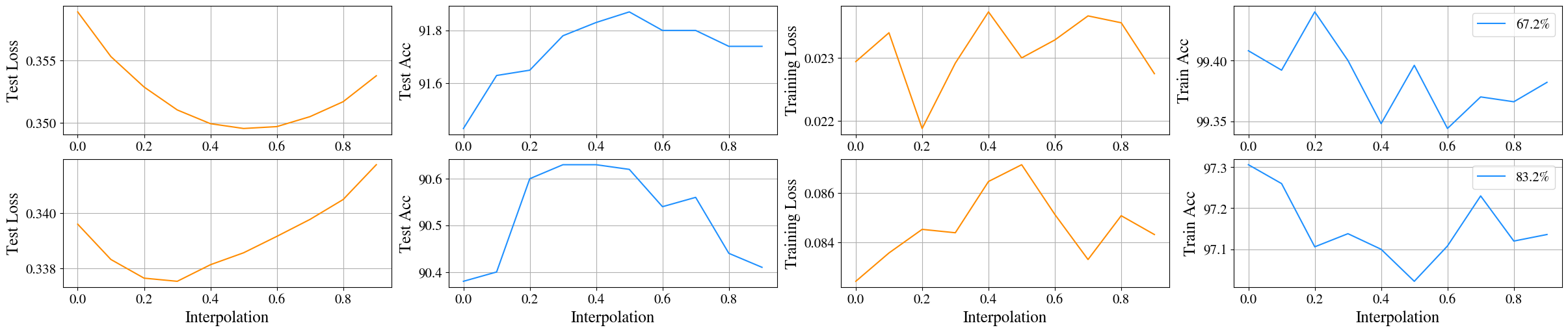}
    \caption{Linear mode connectivity of random networks after repeated cyclic training. Each row corresponds to a sparsity.}
    \label{fig:lmc-2}
\end{figure}

\begin{figure}[h!]
    \centering
    \includegraphics[width=\textwidth]{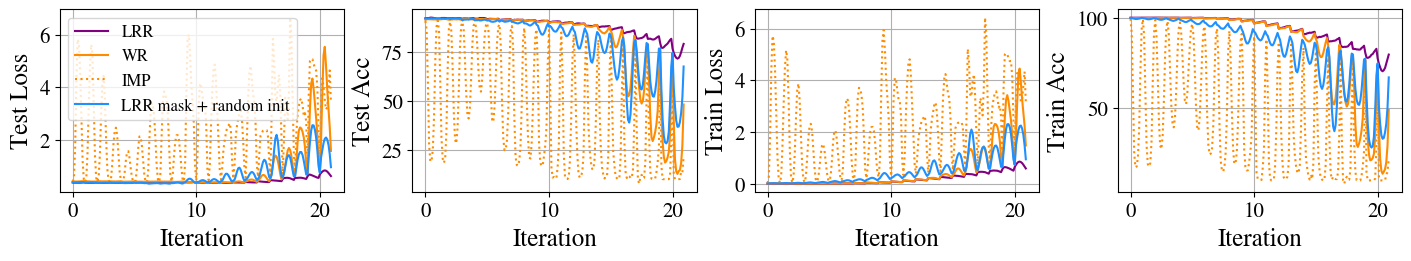}
    \caption{Linear mode connectivity between consecutive masks identified by iterative pruning methods on CIFAR10.}
    \label{fig:lmc-coupling-iterative}
\end{figure}

\begin{figure}[h!]
    \centering
    \includegraphics[width=\textwidth]{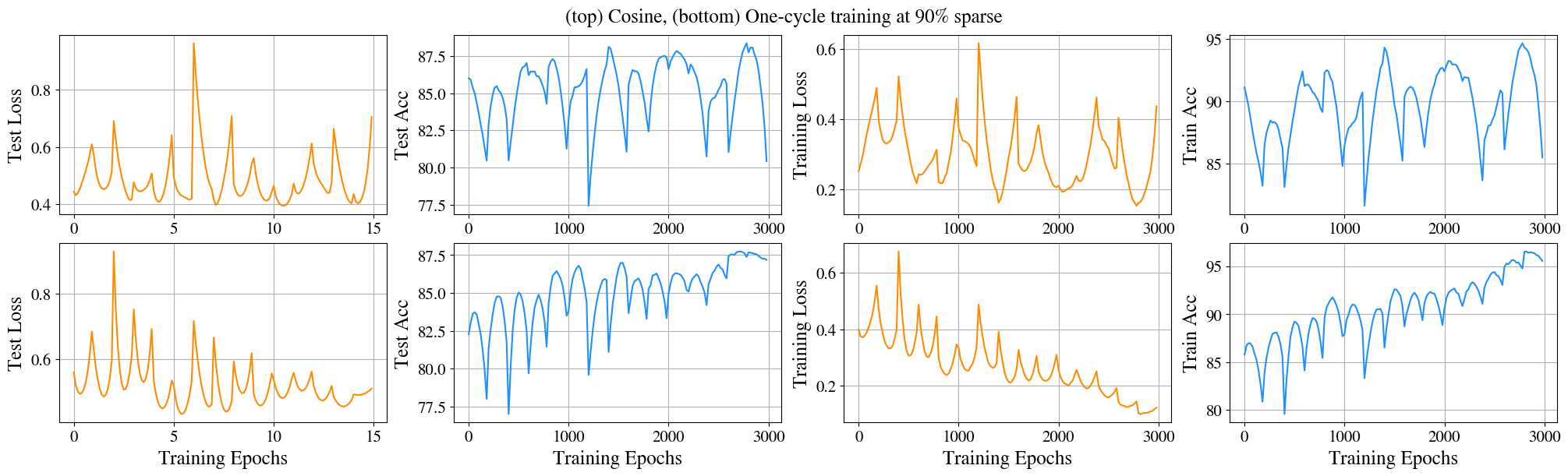}
    \caption{Linear mode connectivity for a  $90\%$ sparse random network with increased training using cosine (top) and one-cycle (bottom) learning rate schedules on CIFAR10.}
    \label{fig:lmc-cosine}
\end{figure}

\begin{figure}[h!]
    \centering
    \includegraphics[width=\textwidth]{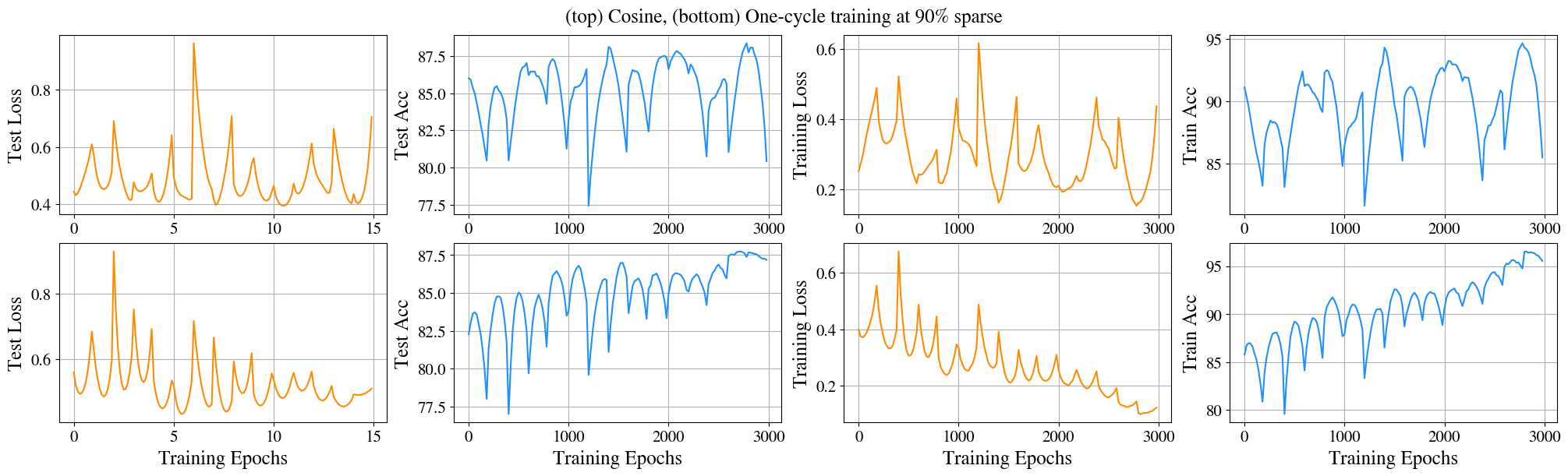}
    \caption{Linear mode connectivity of consecutive training cycles for LRR mask + warmup init (top) and LRR mask + random init (bottom) at $90\%$ sparsity on CIFAR100.}
    \label{fig:c100-lrr-random}
\end{figure}

\begin{itemize}
    \item Figure \ref{fig:lmc-1} shows the connectivity between the first two cycles for cyclic training for a random mask on CIFAR10.
    \item Figure \ref{fig:lmc-2} shows the connectivity between the last two cycles for cyclic training for a random mask on CIFAR10.
    \item Figure \ref{fig:lmc-coupling-iterative} plots the linear mode connectivity of iterative pruning algorithms LRR, WR and IMP as well as an iterative LRR sparse mask with a random init on CIFAR10.
    \item Figure \ref{fig:lmc-cosine} plots the linear mode connectivity for models every $200$ epochs for random sparse networks trained with one-cycle and cosine schedules for $2000$ epochs on CIFAR10.
    \item Figure \ref{fig:c100-lrr-random} shows the linear mode connectivity for a LRR mas + warmup init and LRR mask + random init on CIFAR100 at $90\%$ sparsity to highlihgt the phenomenon of coupling.
\end{itemize}

\end{document}